\documentclass[letterpaper, 10 pt, conference]{ieeeconf}  % Comment this line out if you need a4paper

\IEEEoverridecommandlockouts                              % This command is only needed if
\usepackage{graphicx}      % include this line if your document contains figures
\usepackage{comment}
\usepackage[utf8]{inputenc} 
\usepackage{algorithmic}
\usepackage{graphicx}
\usepackage{textcomp}
\usepackage{xcolor}
\usepackage{subfigure}
\usepackage{cite}
\usepackage{csquotes}
\usepackage{multicol}
\usepackage{tikz}
\usepackage{footnote}
\usepackage{xcolor}
\usepackage{siunitx}
\usepackage{url}

\usepackage{mathtools}

\usepackage{balance}
\usepackage{soul} % Strikethrough text

\usepackage{tikz}

\newcommand\numeq[1]%
  {\stackrel{\scriptscriptstyle(\mkern-1.5mu#1\mkern-1.5mu)}{=}}

\usepackage{hyperref}
\usepackage{times} % assumes new font selection scheme installed
\usepackage{amsmath} % assumes amsmath package installed
\usepackage{amssymb}  % assumes amsmath package installed

\title{\LARGE \bf
    Remote telepresence over large distances via robot avatars: case studies
}

\author{Mohamed Elobaid$^1$,  Stefano Dafarra$^{1}$, Ehsan Ranjbari$^1$,  Giulio Romualdi$^1$, \\ Tomohiro Chaki$^{2}$, Tomohiro Kawakami$^{2}$, Takahide Yoshiike$^{2}$ and Daniele Pucci$^{1,3}$
\thanks{$^{1}$ Artificial and Mechanical Intelligence \emph{AMI} (Italian Insititute of Technology); Genoa, Italy {\tt\small {\{firstname.lastname\}@iit.it}}.}%
\thanks{$^{2}$ Frontier Robotics, Innovative Research Excellence; Honda R\&D,  Saitama, Japan {\tt\small {\{firstname.lastname\}@jp.honda}}}
\thanks{$^{3}$ Machine Learning and Optimisation, The University of Manchester, Manchester, United Kingdom.}%
}

\begin{document}

\maketitle
\thispagestyle{empty}
\pagestyle{empty}

\begin{abstract}     
This paper discusses the necessary considerations and adjustments that allow a recently proposed avatar system architecture to be used with different robotic avatar morphologies (both wheeled and legged robots with various types of hands and kinematic structures) for the purpose of enabling remote (intercontinental) telepresence under communication bandwidth restrictions. The case studies reported involve robots using both position and torque control modes, independently of their software middleware.
\end{abstract}

\smallskip 
%\begin{keywords}

%\end{keywords}
\section{Introduction}
In Walt Disney's \textit{carousel of progress}, actions of an actor wearing a \enquote{control harness} are pre-recorded on tapes and played back on a so-called \enquote{audio-animatronic} figure \cite{disney}. The actions and facial expressions, when played back, are almost life-like inspiring awe in the audience. Even with the lack of feedback to the actor, and the limited distance covered, this represents an early example (being now a 60-year-old attraction) of a telerobotic application. 

Telerobotics and Telepresence applications, however, need not only be social interaction scenarios where maximal human-likeness on the avatar part is beneficial \cite{social_telepresence}, and may require the execution of certain tasks as well as the ability to interact with a possibly hazardous environment and/or humans and other robots \cite{telerobotics}. These utility-driven telepresence applications thus necessitate both manipulation and/or locomotion capabilities on the part of the robotic avatar. On the social, and entertainment side, examples include the wheeled robot winner of the ANA-Xprize telexistence and teleoperation competition held in 2022 \cite{nimbro, ANAAvatar} which was teleoperated by a randomly picked competition judge. On the utility side, an early example is the research on nuclear radioactive materials handling via remotely teleoperated manipulators developed in the first half of the previous century by R.C. Goertz \cite{early_telerobotics}. Other more recent examples include utilizing a VR heat-set and trackers with a Baxter robot (two-arm manipulator) to perform basic manufacturing-like tasks by an operator \cite{baxter}, 3D spray painting with \enquote{immersive} visual feedback via a teleoperated drone \cite{painting_drone}, robot-assisted telesurgery \cite{operation_lind, davinci} and more importantly a humanoid robotic avatar system allowing full embodiment and immersion \cite{icub3_avatar}.

\begin{figure}
    \centering
    \includegraphics[width=0.45\textwidth]{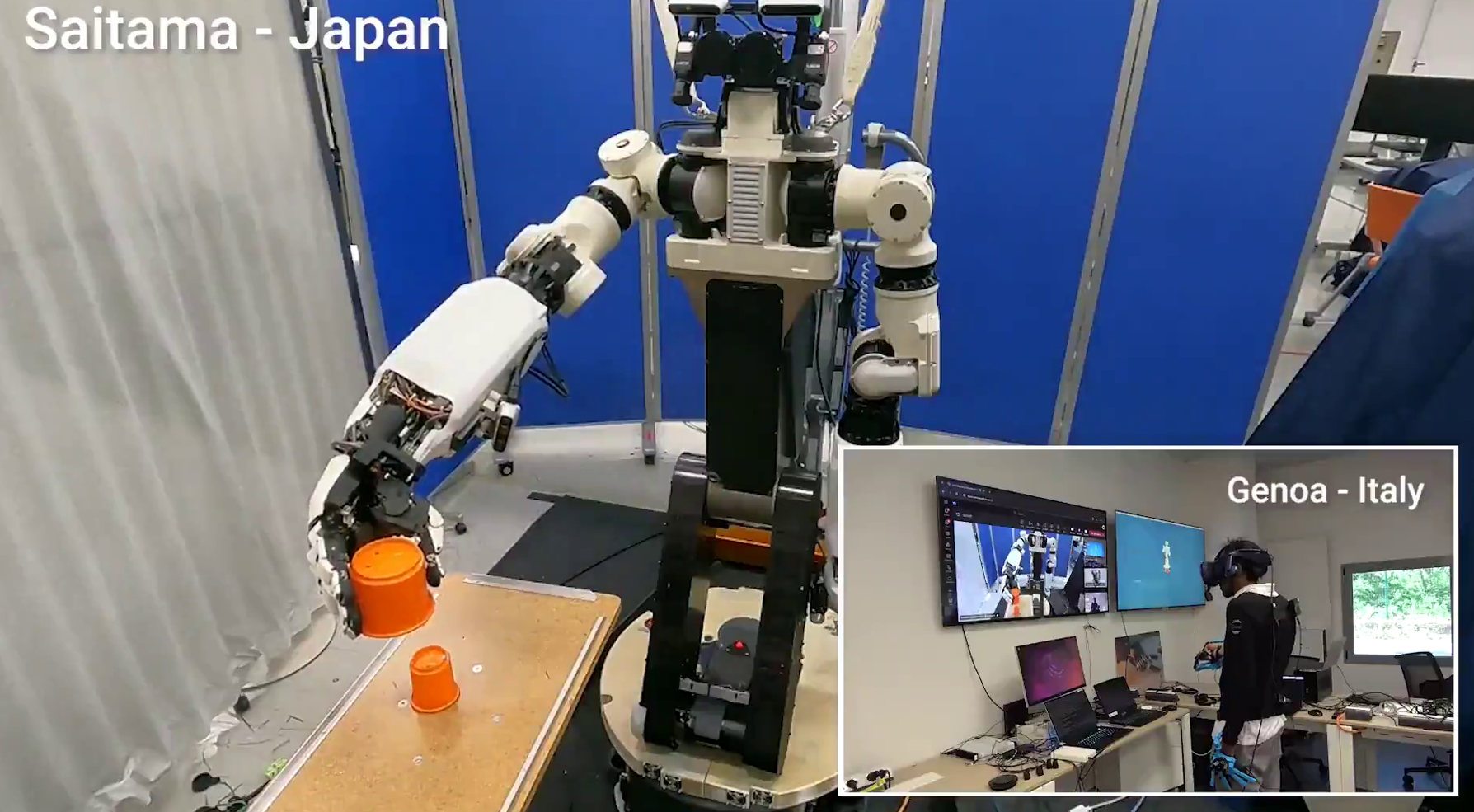}
    \caption{An example of the telepresence avatar system in action with a torque-controlled wheeled robot with a humanoid upper body and a kinematic structure different to that of iCub}
    \label{fig:honda_teleoperation_test}
\end{figure}

From the preceding examples, intuitively it turns out that telepresence through robotic avatars is \enquote{maximized} if; $(i)$ on the operator side; the feeling of being present in a different location to your actual body, with maximal sensation (visual, auditory and touch) and loco-manipulation capability is present, and $(ii)$ on the interaction partner side; verbal and non-verbal social cues are present during an interaction scenario e.g. (gaze and eye-contact, facial expression, gestures, and touch ..etc). These, in turn, necessitate some design choices on the part of the robot avatar. For instance, there needs to be a \enquote{face} encouraging social interaction through mimicking human facial expressions and non-verbal social cues. The hands need to be dexterous facilitating manipulation and grasping. In addition, the robotic avatar needs to have a locomotion mode, enabling the operator to traverse distances remotely.

In a recent work \cite{icub3_avatar}, a fully immersive avatar software and hardware architecture was proposed and validated. While this architecture was designed around the iCub humanoid robot, it was designed to be flexible enough to handle different robot morphologies. In addition, the software components enabling the use of this architecture were made open source.

This paper discusses design choices and necessary adjustments made while leveraging the above-mentioned architecture for the purpose of enabling telepresence in different settings using different robots over very large distances. Namely, the robot avatars in question will have different morphologies (upper body kinematic structures to those in \cite{icub3_avatar}), different locomotion interfaces and different hands structures and fingers coupling laws.

The above discussion will be carried over the following case studies; $(i)$ the remote teleoperation of the ergoCub humanoid robot between Italy on the one hand, and London, Bruxelles on the other, and $(ii)$ the teleoperation of a wheeled torque-controlled robotic avatar between Italy and Japan. 

This paper is organized as follows: Section \ref{sec:two} presents the case where the ergoCub humanoid is leveraged as an Avatar to attend the IEEE ICRA 2023 remotely from Italy as well as later visit the European - EU -  Parliament. Section \ref{sec:three} will present the main case study, in which a 'wheeled robotic avatar' is being teleoperated intercontinentally. Section \ref{Sec:four} presents shortcomings and lessons learned.  Concluding remarks in Section \ref{sec:conclusion} end the manuscript. 

\section{Participating in an International conference and visiting the EU parliament  via a humanoid avatar}\label{sec:two}

In this section, we present results pertaining to the use of the architecture presented in \cite{icub3_avatar} with the recently introduced humanoid robot ergoCub \cite{ergo_cub}. The main novelty here is the use of hardware accelarated compression and haptics message structure adjustments to reduce bandwidth requirements allowing for better performances. This aspect is made evident when comparing the first small case study of participating in the IEEE ICRA 2023 conference held in London by an operator located in Genoa, compared to the more recent visit of the robot to the EU parliament in Brussels by an operator in Genoa. In the latter case, both bandwidth requirements by the gloves and that of the visual feedback are largely reduced leading to a more immersive telepresence experience.

\subsection{Remote telepresence bandwidth requirements}\label{bandwidthsec}

The architecture presented in \cite{icub3_avatar} typically requires more than $\SI{20}{mbits/s}$ bandwidth. The main culprits are the visual feedback where no compression and image preprocessing techniques were utilized, and the haptic gloves message frequency and structure. A single image during the XPrize participation detailed in \cite{icub3_avatar} required $\SI{8}{mbits/s}$ bandwidth, where in that case the robot had two cameras one for each eye. Similarly, the gloves message (a thrift containing joint states, force and vibro-tactile feedback) required around $\SI{10}{mbits/s}$ running at $\SI{100}{Hz}$. These aspects hindered a smooth telepresence experience on the operator part. In fact, when network resources available at a given venue are limited, the operator would have to deal with siginificant delays both on the camera stream and on commanding a specific motion on the robot, producing high cognitive load. Motivated by this discussion,  the developments in two directions were carried out, namely;
\begin{itemize}
    \item Reduction of the bandwidth required by the gloves commands for fingers' motion retargeting in grasping applications. This was done by first leveraging the YARP-based wrapper/remapper architecture where a  lower frequency of $\SI{10}{Hz}$ was used to get data from the physical glove device (in this instance the SenseGlove DK1) while the control boards' frequency is left at higher rates. Second, by combining the messages of all fingers for both joint state values and haptic feedback in a single custom-made thrift message type.
    \item The use of hardware-accelerated compression of the camera feedback. This aspect is incorporated in the official latest planned release of YARP (version $3.10$), enabling the possibility to use all the encoders and decoders available with ffmpeg \cite{ffmpeg}. This development enabled a reduction of the required bandwidth for single images to roughly $\SI{1.5}{mbit/s}$. %Better performance is also being investigated. It is however important to stress that this development was not utilized in this telepresence scenario and we rather opted to utilize a separate link for visual feedback.
\end{itemize}

\begin{figure}
    \centering
    \includegraphics[width=0.47\textwidth]{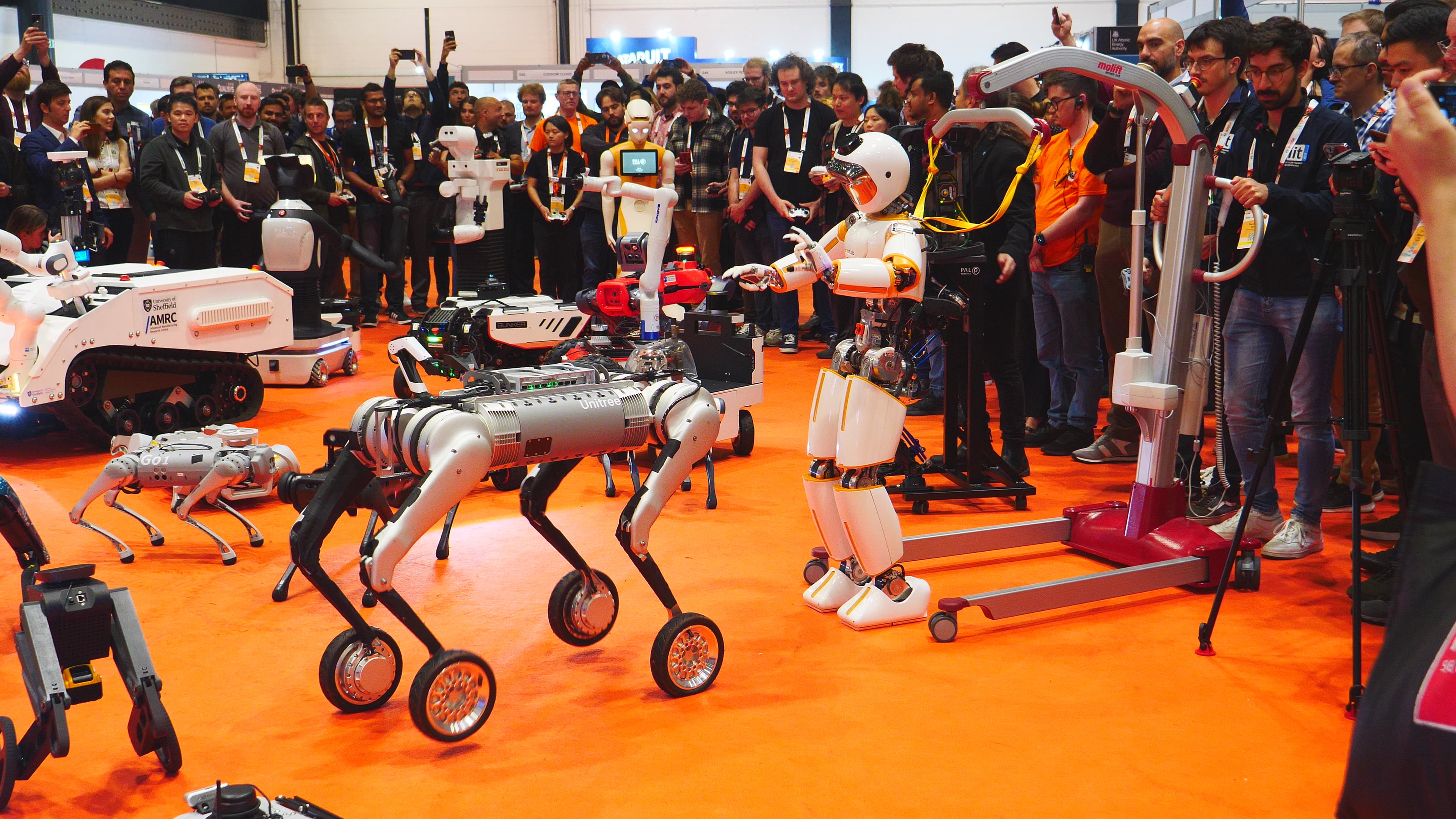}
    \caption{The ergoCub robot interacting with other robots in London during ICRA 2023 while being teleoperated from Genoa.}
    \label{fig:icra2023}
\end{figure}

\begin{figure}
    \centering
    \includegraphics[width=0.47\textwidth]{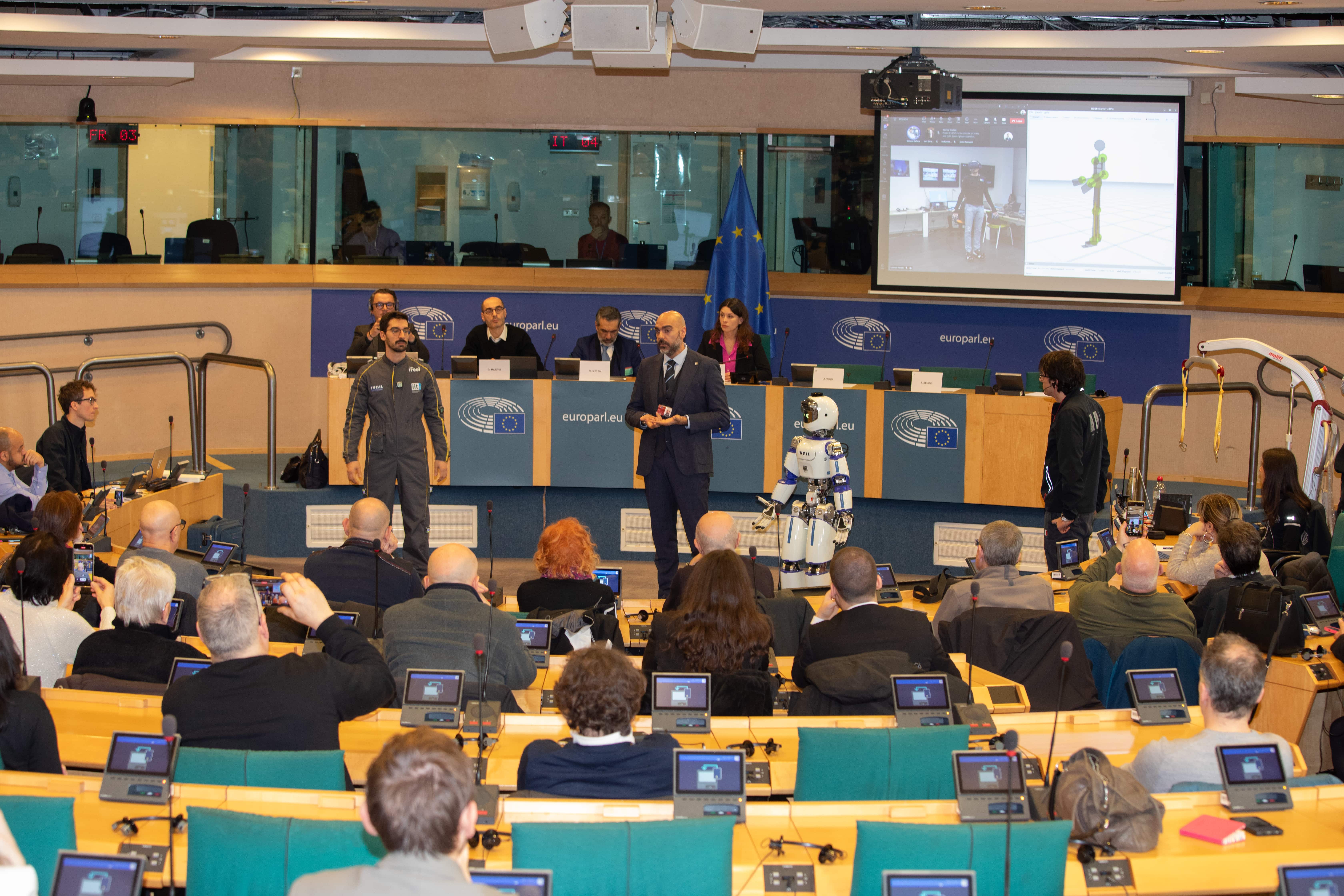}
    \caption{During a talk in the European Parliament in Brussels, a remote operator pictured in the large screen is listening attentively through the ergoCub avatar.}
    \label{fig:eu_parliament}
\end{figure}

\subsection{Participating in ICRA 2023 remotely}

Over the span of 4 days, between Monday 29th May 2023 and Thursday 1st of June 2023, several operators interchanged on commanding the robot located on the ICRA exhibition hall in London from Genoa.

The most pertinent aspects of this experience are twofold. First, the fact that among several teleoperated robots in the arena, ergoCub was the only robot being teleoperated from a different city/country.  Second, no bandwidth requirement reduction in terms of image compression nor gloves feedback frequency adjustment was applied. The robot was allowed to roam the exhibition hall, controlled by the remote operator utilizing the iFeel Walking Retargeting locomotion interface discussed in \cite{icub3_avatar}. The result while important, being, to the best of the authors knowledge,  the first time a remote international participation of this kind to a major robotics conference was carried out, was not immersive enough for a full telepresence experience (see Figure \ref{fig:icra2023}). The operator needed to adjust to significant delays as well as some sudden small lapses of visual feedback loss. One interesting observation is that some of the audience in London did not realize that the robot was remotely teleoperated from Genoa when interacting with it until they were informed of that fact. This highlights that while on the operator part, the telepresence experience was not \enquote{optimal}, on the interaction partner part, the social aspects were enough to derive an immersive interaction.

\subsection{Visiting the EU Parliament remotely}

Having specified bandwidth requirements as the major blocking point, on February 2024 a visit to the European Parliament in Brussels where a session on \enquote{Robotics and AI in the European Landscape} was held, proved to be an excellent moment to test the technical adjustments discussed above in subsection \ref{bandwidthsec}.

The result achieved was an almost latency-free telepresence experiment, where the total bandwidth required for running the whole-body retargeting and walking was reduced from almost $\SI{30}{mbits/s}$ to around $\SI{15.5}{mbits/s}$. %This experiment also entailed interacting with persons on the stage, and demonstrating the ability to carry payloads and deliver them in an interactive manner. 

\section{On the teleoperation of a torque-controlled robot between Italy and Japan}\label{sec:three}

\begin{figure*}
    \centering
    \includegraphics[width=\textwidth]{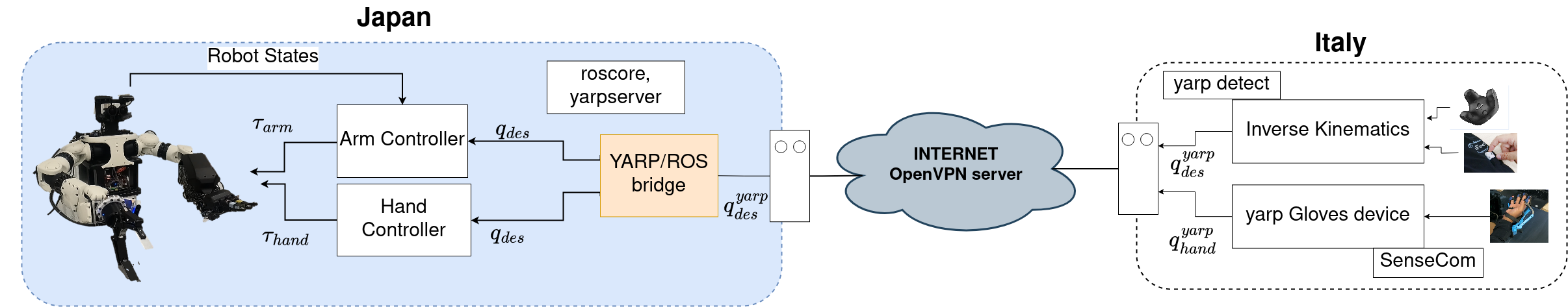}
    \caption{Overview of the architecture enabling the teleoperation of the Honda avatar robot in Japan from Genoa}
    \label{fig:honda_case_study_architecture}
\end{figure*}

Our main case study serves to highlight different aspects and design choices necessitated by both the robot hardware and its software architecture. The results presented in this section were made possible due to a collaboration between the Artificial and Mechanical Intelligence research line in Italy and the Innovative Research Excellence unit of the Honda R\&D department in Japan within the context of a joint research project. The overall objective of this activity is the realization of a \enquote{teleoperation and control architecture for collaborative payloads handling in a shared autonomy manner} for Honda-produced robotic avatars \footnote{the reader may check \url{https://global.honda/en/tech/Avatar_robot}}. To make the discussion concrete, we will make reference to Figure \ref{fig:honda_case_study_architecture}. The case study will report experimental activities concerning the control of the upper-body (mainly arm and hands) remotely. Consideration for the mobile base locomotion interface will be given in later sections through simulations. 

\subsection{The Honda avatar robot}

The current version of the Honda avatar robot is a wheeled robot with a humanoid-like upper body.  The robot stands at $\SI{155}{cm}$ tall and weighs $\SI{101}{kg}$. It comprises 39 joints, of which 34 are controllable axis, and a mobile base with four passive omni-wheel mechanisms and two driving wheels. The robot head is equipped with a multi-sensor wide-view fisheye and 4-RGBD camera, together with Intel realSense \cite{realsense} cameras at the hand. The robot is torque-controlled with a proprietary whole-body optimization-based torque controller that considers self-collisions and joints and motors' current limits \cite{honda_qp_icra}. 

\begin{figure}[h]
    \centering   \includegraphics[height=0.45\textwidth]{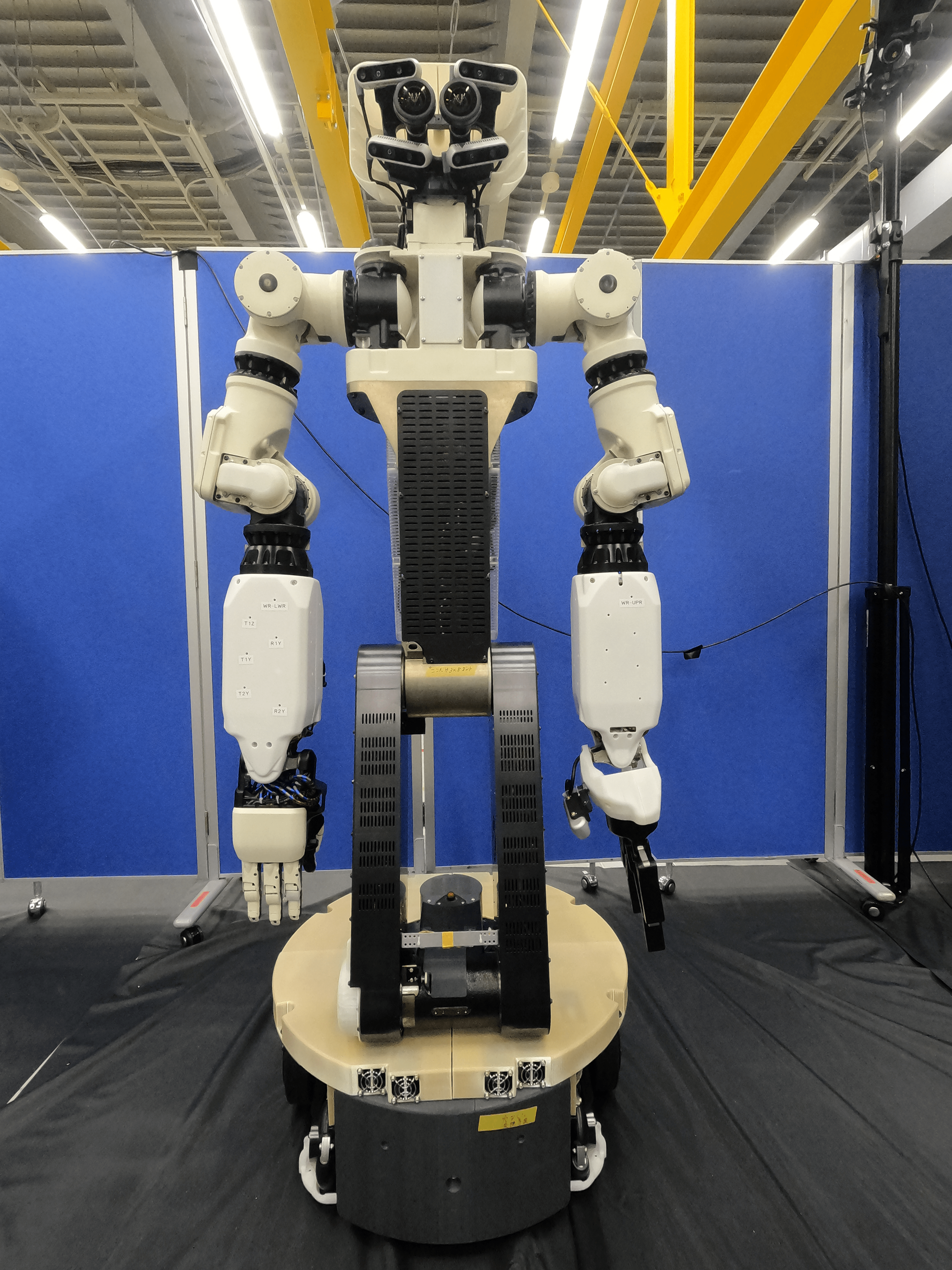}
    \caption{The Honda avatar robot}
    \label{fig:horst_avatar}
\end{figure}

As can be seen from Figure \ref{fig:horst_avatar}, this avatar robot already presents a different morphology to the iCub3 humanoid over which the avatar architecture presented in \cite{icub3_avatar} was validated.

\subsection{The dexterous Honda hand}
The advanced Honda hand \cite{hmf_hand} comprises a forearm and a 4-fingered hand. The hand and wrist have in total 18 joints of which 13 are \enquote{drive joints}. The hand palm has 2 cameras and an additional camera is attached to the forearm. Those cameras are part of an AI-assisted telepresence architecture that allows refining the commanded motions. These aspects are outside the scope of this work. The hand is able to produce a maximum grasp force of $\SI{50}{N}$, and is equipped with 4 6-axis force/torque sensors. 

It is rather clear that the number of fingers and the proprietary associated coupling laws between the drive and interlocked joints are substantially different to those present in iCub3 (and as we will see later ergoCub), requiring some adjustment on the implementation of the control commands of the fingers. To this end, a specific hand control coupling handler class was implemented to map actuation commands to fingers joints.

\subsection{The software architecture}

The software driving and controlling the robot is comprised of ROS \cite{ros} nodes performing hands retargeting and inverse kinematics computations running at a lower frequency (around $\SI{25}{Hz}$) and an Embedded components middleware responsible for the whole-body control, joints control, and torque-based stabilization running at a higher frequency (around $\SI{500}{Hz}$). The joint controller expects a ROS joint states message comprising the joints of the dexterous Honda multi-finger hand while the stabilizer expects references on joint positions, velocities, and accelerations for the robot body.

Compared to the avatar architecture presented in \cite{icub3_avatar}, the above software architecture presents some technical interfacing challenges that were taken into consideration when teleoperating the robot. Those can be summarized as 
\begin{itemize}
    \item \textit{\textbf{Middle-ware interface}}: commands generated in Genoa with YARP \cite{yarp} based software needed to be interpreted as the appropriate ROS message types. Several options are available to carry out this task. For instance, YARP devices, when implemented could be tailored to publish ROS message types. A simpler and more direct approach is to design a thin interface layer, herein called a yarp-ros-bridge \footnote{the implementation can be found in \url{https://github.com/ami-iit/yarp-ros-bridge/}}. This bridge, running on callbacks to minimize latency, allows a user to specify, using a configuration file, a pair of a ROS topic and a YARP port, the direction of the message being transmitted and the message type. For the purpose of the case study, only the required types of messages were supported keeping the implementation compact.
    \item \textit{\textbf{Division of labor}}: the role of the inverse kinematics and human hand pose retargeting components are given to the YARP based software modules detailed in \cite{icub3_avatar}. For the inverse kinematics computations, the model of the robot is taken into consideration. This entailed performing some computations where we aligned the frames of the robot links to those of the sensors used to specify the inverse kinematics tasks. In this setting, the inverse kinematics comprised of Cartesian tasks (i.e. link poses) given by the HTC Vive trackers \cite{vive} and gravity tasks aligning the gravity vector of specific links given by measurements of the iFeel nodes \cite{ifeel}. 
\end{itemize}

\subsection{The telepresence scenario}

The scenario envisioned is a puzzle-like one where a small bucket is placed on a table in front of the robotic avatar in Japan. The operator in Genoa should move the upper-body/arms to grasp the bucket utilizing the fingers. An interaction partner in Japan then places a smaller bucket on the table and the operator is asked to place the bigger bucket on top of the smaller one. To do this, the operator relies mainly on the arms and dexterous hands.

\subsection{Results}
When setting up the communication network configuration, Honda opted for a limited bandwidth Star Link\copyright \cite{starlink}. In the final demonstration, the bandwidth available was around $\SI{300}{kbits/s}$ in both directions. Compared to the usual optimized bandwidth requirement of $\SI{15.5}{mbits/s}$ when running the avatar infrastructure, this posed a significant challenge. This aspect, compounded with the extreme distance over which the telepresence scenario was carried introduced large delays (almost $\SI{1}{s}$) for which the operator needed to adjust. Note that light would require $\SI{30}{ms}$ to travel the same distances, and usual pings are typically in the order of $\SI{100}{ms}$ using a fast Internet link between Japan and Genoa. %These aspects motivated the developments in two directions with which one reduces the bandwidth requirements of the original avatar architecture presented in \cite{icub3_avatar}, namely

\begin{comment}
\begin{itemize}
    \item Reduction of the bandwidth required by the gloves commands for fingers' motion retargeting in grasping applications. This was done by first leveraging the YARP-based wrapper/remapper architecture where a  lower frequency of $\SI{10}{Hz}$ was used to get data from the physical glove device (in this instance the SenseGlove DK1) while the control boards' frequency is left at higher rates. Second, by combining the messages of all fingers for both joint state values and haptic feedback in a single custom-made message type, termed a \textit{wearable message}. These modifications were utilized during the experiment reported above.
    \item The use of hardware-accelerated compression of the camera feedback. This aspect is incorporated in the official latest planned release of YARP (version $3.10$), enabling the possibility to use all the encoders and decoders available with ffmpeg \cite{ffmpeg}. This development enabled a reduction of the required bandwidth for single images from $\SI{8}{mbit/s}$ to roughly $\SI{1.5}{mbit/s}$. Better performance is also being investigated. It is however important to stress that this development was not utilized in this telepresence scenario and we rather opted to utilize a separate link for visual feedback.
\end{itemize}
\end{comment}

To that end, the visual feedback of the fisheye camera was done over a separate link (MS teams), while the modifications pertaining to the message of the gloves were applied having both the data from the physical glove as well as the control boards frequency running at $\SI{10}{Hz}$ for maximum bandwidth requirement reduction. With the above setup, the operator was able to adjust to the cognitive load required and execute the puzzle scenario while communicating with the interaction partners on the Honda side. %Media files of the experiments together with relevant data can be found in ... ( Figure \Moesays{remember to add link} \ref{fig:honda_teleoperation_test} is a snapshot). 

\section{General considerations and discussion}\label{Sec:four}

In this section were provide some general comments and lessons learned from the above experiments.

\subsection{Comments on the locomotion interface}

While the telepresence case study of the Honda avatar robot described above does not entail any locomotion task, the overall objectives of that particular project necessitate the extension of the locomotion interfaces presented in \cite{icub3_avatar} to handle wheeled robots. To this end, a locomotion interface utilizing the HTC Vive trackers on the waist and feet of a (possible) seated operator was developed. This interface generates a triplet of velocities being linear, angular, and lateral velocities respectively. The operator commands the velocities by putting forth one foot beyond an \enquote{idle area} in a specific direction, or rotating in place while keeping the feet inside the idle area. This uniform interface allows to command of both wheeled robots (with a differential drive mobile base like the R1 robot \cite{r1_robot} shown in simulations, or even  omni-wheeled mobile bases) and bipedal humanoids using a unicycle footsteps planner (e.g. iCub3 and ergoCub).

\begin{figure}[h]
    \centering
    \includegraphics[width=0.4\textwidth]{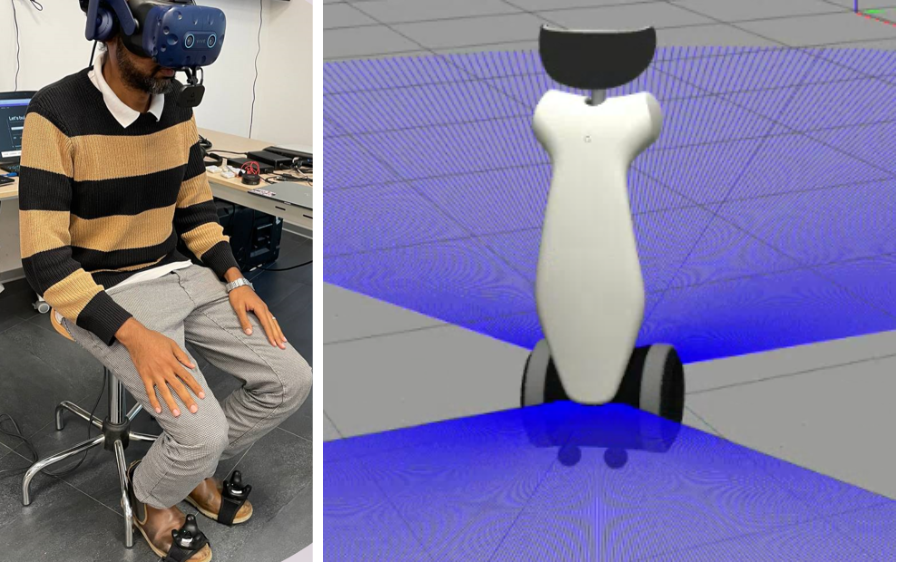}
    \caption{From left to right; a seated operator wearing the HTC Vive headset and trackers to command velocities of the R1 robot with a differential drive mobile base.} %On the right is the Honda Avatar robot omni-wheeled mobile base.}
    \label{fig:enter-label}
\end{figure}

\subsection{Comments on the visual feedback}

\begin{figure}[h]
    \centering
    \includegraphics[width=0.46\textwidth]{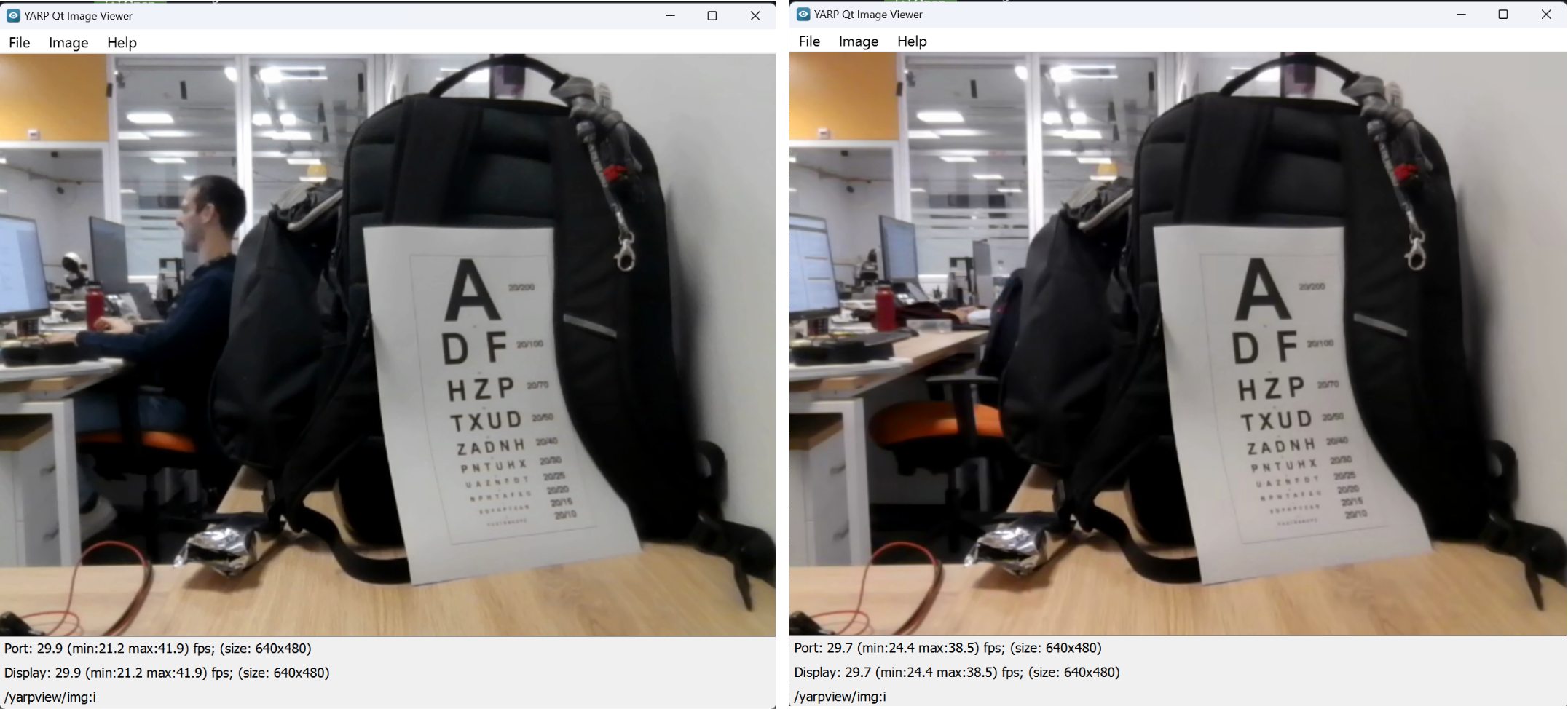}
    \caption{Quality vs bandwidth with hardware accelerated encoding via ffmpeg. On the left the image requires $\SI{7.5}{mbits/s}$, while  on the right requires only $\SI{800}{kbits/s}$.}
    \label{fig:ffmpeg}
\end{figure}

The main immersion breaking bottle-neck in the previous case studies has to do with visual feedback. On the one hand, the limited field of view of the camera hindered the operator's ability to fully embody the robotic avatar in the ergoCub case. On the other hand, the large latency and lower quality/resolution put a high cognitive load on the part of the operator. Concerning the former, different rendering techniques coupled with a wider-angle camera are being investigated. While for the latter, hardware compression techniques proved very reliable as attested by the last case study (see Figure \ref{fig:ffmpeg}).

\section{Conclusion}\label{sec:conclusion}

Several case studies on telepresence via robotic avatars with different morphology over large distances were presented utilizing a recently proposed flexible architecture. The adjustments and modifications allowing such flexibility are made concrete, and shortcomings are put in light. Future work concerns on the one hand the study of social telexistance scenarios and the impact of such technology, and on the other hand, further improvements in considering the effect of inherent latency as part of the design together with improvements on the visual and force feedback for better immersion.

\section*{Acknowledgment}
This work was partially supported by the Italian National Institute for Insurance against Accidents at Work (INAIL) ergoCub Project, and Honda Research and Development Japan, through a joint-lab research initiative.

%\balance
\bibliographystyle{IEEEtran}      
\bibliography{biblio}                  

\end{document}